%% file: main.tex
\documentclass{article}
\usepackage{spconf,amsmath,amssymb, booktabs}
\usepackage[pdftex]{graphicx}
\usepackage{bbm}
\usepackage{multirow}
\usepackage{hyperref}

\usepackage{url}
\usepackage{hyperref}
\hypersetup{colorlinks=True, urlcolor=black}

\def\FFM{{\texttt{FFM}}}
\def\NCB{{\texttt{NCB}}}
\def\TB{{\texttt{TB}}}

\newcommand{\tworow}[1]{\multirow{2}{*}{\centering #1}}

\usepackage{xcolor}
\usepackage{tipa}

\newcommand{\expID}[1]{$\mathrm{#1}$}

\title{Practical Conformer: Optimizing Size, Speed and Flops \\ of Conformer for On-Device and Cloud ASR}
\name{
\begin{tabular}{c} Rami Botros\textsuperscript{*}, Anmol Gulati\sthanks{Equal contribution}, Tara N. Sainath, Krzysztof Choromanski,  \\ 
Ruoming Pang, Trevor Strohman, Weiran Wang, Jiahui Yu
\end{tabular}
}
\address{Google LLC, USA \\
\fontsize{9}{9}\selectfont\ttfamily\upshape
\{ramibotros, anmolgulati, tsainath, kchoro, rpang, strohman,weiranwang,jiahuiyu\}@google.com}
\begin{document}
\ninept
\maketitle
\newcommand{\TDS}[1]{\textcolor{red}{strohman:#1}}
\input{abstract}
\input{introduction}
\input{modeling}
\input{experiments}
\input{results}
\input{conclusions}

\bibliographystyle{IEEEbib}

\bibliography{ref}
\end{document}

%% file: abstract.tex
\begin{abstract}
Conformer models maintain a large number of internal states, the vast majority of which are associated with self-attention layers. With limited memory bandwidth, reading these from memory at each inference step can slow down inference. In this paper, we design an optimized conformer that is small enough to meet on-device restrictions and has fast inference on TPUs. We  explore various ideas to improve the execution speed, including replacing lower conformer blocks with convolution-only blocks, strategically downsizing the architecture, and utilizing an RNNAttention-Performer. Our optimized conformer can be readily incorporated into a cascaded-encoder \cite{arun21cascade} setting, allowing a second-pass decoder to operate on its output and improve the accuracy whenever more resources are available. Altogether, we find that these optimizations can reduce latency by a factor of 6.8x, and come at a reasonable trade-off in quality. With the cascaded second-pass, we show that the recognition accuracy is completely recoverable. Thus, our proposed encoder can double as a strong standalone encoder in on device, and as the first part of a high-performance ASR pipeline.
\end{abstract}

%% file: introduction.tex
\section{Introduction} \label{sec:intro}

End-to-end (E2E) ASR models, which combine acoustic, pronunciation and language models from conventional systems \cite{Golan16} into one neural network, have become an active research area in the past few years \cite{bo21system,Ryan19,CC18,KimHoriWatanabe17,JinyuLi2019,Zeyer2020}. Since they are a fraction of the size of conventional models, their inference speed is often much faster \cite{bo21system,Ryan19,sainath2021cascadedlm,sainath2020streaming}, which makes them attractive for various live applications.

There is considerable interest to further improve inference speed of E2E models, and to better utilize large cores on Tensor Processing Unit (TPU) devices in particular. Some works have replaced the LSTM encoder and decoder of these models with parallelizable networks. For example, \cite{Rami21} looks at improving speed of the E2E decoder by replacing a large LSTM with a simple embedding decoder. On the encoder size, transformer \cite{Zhang20,tsunoo2020streaming} and conformer \cite{gulati2020conformer} architectures, which have no recurrent connections, enable batching across multiple frames. While specialized hardware, such as on-device edge TPUs, significantly speed up computation on a single utterance, cloud TPUs, which process requests from numerous users, have additional bandwidth constraints with conformers~\cite{JouppiYoungPatilEtAl17}. 

A major issue with the conformer encoder is that the number of internal states to maintain is much larger than in the LSTM case. Most of these correspond to the \textit{key} and \textit{value} tensors in self-attention. While TPUs certainly help with parallelized computation of attention, inference is often still slowed by the memory-bandwidth cost of repeatedly loading such states \cite{shazeer2019fast}. For many streaming applications, in order to display the words as soon as the model can output them, it is often preferable to forgo batching across frames. Instead, the encoder is given just a few input frames at a time. This forces frequent reads of the state vectors, which exacerbates the memory-bottleneck issue and undermines the goal of fast outputs. In practice, our benchmarks have show that switching from LSTM to conformer encoder increases the per-step inference latency 10x.

In this work, we seek to design a conformer that can be used as a streaming, causal encoder for different TPU environments, such as cloud-based TPUs, as well as on-device edge TPUs. To qualify as low-cost for both settings simultaneously, the design needs to meet a compounded set of conditions. Accordingly, we look for architectures that meet realistic criteria with respect to: limited model size, TPU-latency and number of floating-point operations (flops). Specifically, we seek a solution where the cloud-TPU latency is below 5ms, with flops below 100M and a model size below 50M. We include a restriction on flops, since it has been suggested that they can be predictive of energy consumption \cite{tang2018flops,tang2018experimental} ---~something that our earlier experiments have confirmed. Within these bounds, we try to obtain the best possible accuracy.

Moreover, we seek a solution where the output of our optimized conformer should successfully serve two distinct types of downstream networks: (1) An RNN-T decoder directly, or (2) a second (cascaded) encoder whenever the computational resources and/or latency requirements can permit it. The first works much closer to the word-level modality, and can typically have a very different architecture from the second. Hence, it is reasonable to expect both downstream networks to require different types of information and representations from their inputs. Thus, achieving strong performance inside as well as outside the cascade setting can be seen as a multi-task target for our optimized encoder, and can become more challenging as we limit its capacity by making it smaller. 

To achieve our goals, we first look at replacing the lowest blocks of conformer with convolution-only blocks which do not suffer from large-state issues. In addition, we strategically downsize some of our model's dimensions to reduce size and computation. Finally, we explore the use of performer \cite{performers} as a way to improve conformer speed by avoiding explicit materialization of the attention tensor.

There are many methods that improve the computational efficiency of self-attention, as summarized in \cite{emformer,fastformer}. As with \cite{performer_in_conformer}, we opt for performer layers \cite{performers} inside the conformer, which can be used as a drop-in replacement for self-attention and capture long-context dependencies. Unlike performer, emformer from \cite{emformer} chunks the utterance into segments and parallelizes computation within them. Due to the importance of immediate textual for our application, we do not compare against such techniques.

Another example for speeding up conformer specifically is \cite{burchi2021efficient}, which progressively downsamples the signal along the time dimension. As with our research, the work targets the first few conformer blocks, which precede the downsampling and tend to be the most expensive. For that, the authors use so-called grouped attention, which reduces attention complexity by reshaping its input, stacking neighboring frames together into the depth dimension. In our case, we find that the self-attention layers can be removed from the earliest blocks altogether instead. For higher blocks in our model, we replace self-attention with performer layers, which lowers the cost from quadratic to linear in the length of the input sequence.

Another direction in the literature works on reducing overall model size of the network. Dynamic sparsity \cite{Wu21} is one example, but its sparse operations are not supported on all TPUs, so we do not consider it here. As an alternative, motivated by works such as \cite{zhang2021beyond}, we explore statically removing some connections from our fully-connected layers. Our early experiments, which we omit here for brevity, showed that this works well with our 100M-parameter models, but causes substantial quality degradation for smaller ones.

Distillation with RNN-T from a large to small model has also been investigated~\cite{panchi2020distillation}. This often requires a large teacher model to be trained independently. Recently, in-place distillation \cite{yu21universalasr} addresses this by jointly training a full-context teacher, while distilling it into a smaller streaming student, all as part of a single model. We explore this as part of our cascaded encoder setting, distilling the output of a non-causal 2nd-pass to a smaller causal 1st-pass. 

We investigate the proposed ideas on a large-scale Voice Search task. In total, with our three proposed changes, we are able to compress our original 120M parameter model by around 2.1x in size and 2.7x in flops, with a 6.8x speedup on cloud TPU. While these optimizations give a relative WER degradation of 16\%, we show that if the environment permits more relaxed constraints, cascading a 2nd-pass encoder can fully recover the accuracy without requiring any change in our optimized first-pass design. Hence, our small conformer can be trained to do two jobs simultaneously: Produce a fast output in a limited-resource setting, and potentially pass a useful representation to a larger second-pass in a high-resource setting.

%% file: modeling.tex
\section{Modeling} \label{sec:model}

\subsection{Baseline Conformer}

\begin{figure}[h]
  \centering
  \includegraphics[width=0.49\textwidth]{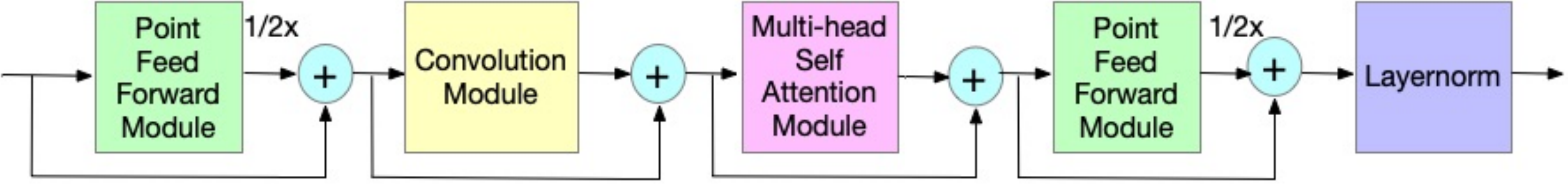}
  \caption{\small{Illustration of conformer block \cite{gulati2020conformer}. }}
  \label{fig:conformer}
\end{figure}

Our baseline conformer encoder \cite{gulati2020conformer} consists of 12 conformer blocks, shown in Figure \ref{fig:conformer}. Each block comprises a stack of 5 modules. First,  a feed-forward module projects the input features to a larger dimension by a factor $FFM$, followed by a non-linear activation, then another linear layer to project the features back to their original dimensions. Next, a convolution module aggregates information from neighboring context to capture relative-offset-based local interactions. Then, a self-attention module allows the model to look back $L$ previous frames, and converts this into a fixed-length vector, capturing more global patterns. Afterwards, another feed-forward module operates on the output of the self-attention module. Finally, a layernorm module helps improve quality \cite{bo21system}.

\cite{shazeer2019fast} describes a source of slowdown for the self-attention module. Specifically, incremental inference (when parallelization is not possible) is often slow due to the memory-bandwidth cost of repeatedly loading in large key and value tensors, which we will call \emph{states}. For example, an 8-layer LSTM with 640 dimensions/layer has roughly 5,120 states per frame. In comparison, a conformer encoder that has 12 layers, 23 frames of left context per layer and is 512 dimensions/layer has roughly 141,312 states, almost 30x as much.

To quantitatively benchmark the latency issue with conformer, Table \ref{tab:lstm_vs_conformer} shows the WER and average per-frame TPU latency of the LSTM \cite{Ryan19} and conformer encoders with the parameters described above. While the two models are roughly equal in size, conformer has much larger latency and flops. In the rest of this section, we describe techniques used to bring down these costs.

\renewcommand{\arraystretch}{0.97}
\begin{table}[h!]
    \centering
    \caption{WER and Latency of 1st-pass encoders}
    \begin{tabular}{lcccc} 
    \toprule
    Encoder & WER & Size & TPU-latency & Flops \\
    \midrule
     LSTM & 6.8  &  110M & 2.4 ms & 198M \\ 
     Conformer & 6.5  & 120M & \textbf{21.8 ms} & \textbf{247M} \\
     \bottomrule 
    \end{tabular}
    \label{tab:lstm_vs_conformer}
\end{table}

\subsection{Removing self-attention layers}

First, we hypothesize that low-level features learned by the first few conformer blocks can be captured by simpler layers. Thus, to reduce computation and size, we remove self-attention layers from the lowest blocks, relying just on the convolution and feed-forward modules in Figure \ref{fig:conformer}. This frees some space and compute resources to make other layers deeper and wider, as discussed in the next section. 

\subsection{Strategic Resizing}
\label{subsection:resizing}
Given the approximate constraints discussed in Section \ref{sec:intro}, specifically with TPU-latency less than around 5ms and flops below 100M, we carefully chose the widths and depths of our layers to get the highest accuracy from a 50M-parameter encoder. While conforming to our size requirement, we focus on tuning 3 hyperparameters of our model's architecture. The FF-module expansion factor (\FFM), the number of convolution-only blocks at the bottom (\NCB), and the total number of conformer blocks (\TB). We will show an ablation study varying these parameters in Section \ref{sec:results}.

\subsection{Implicit Attention with RNNAttention-Performers}

\begin{figure}[t]
  \centering
  \includegraphics[width=0.49\textwidth]{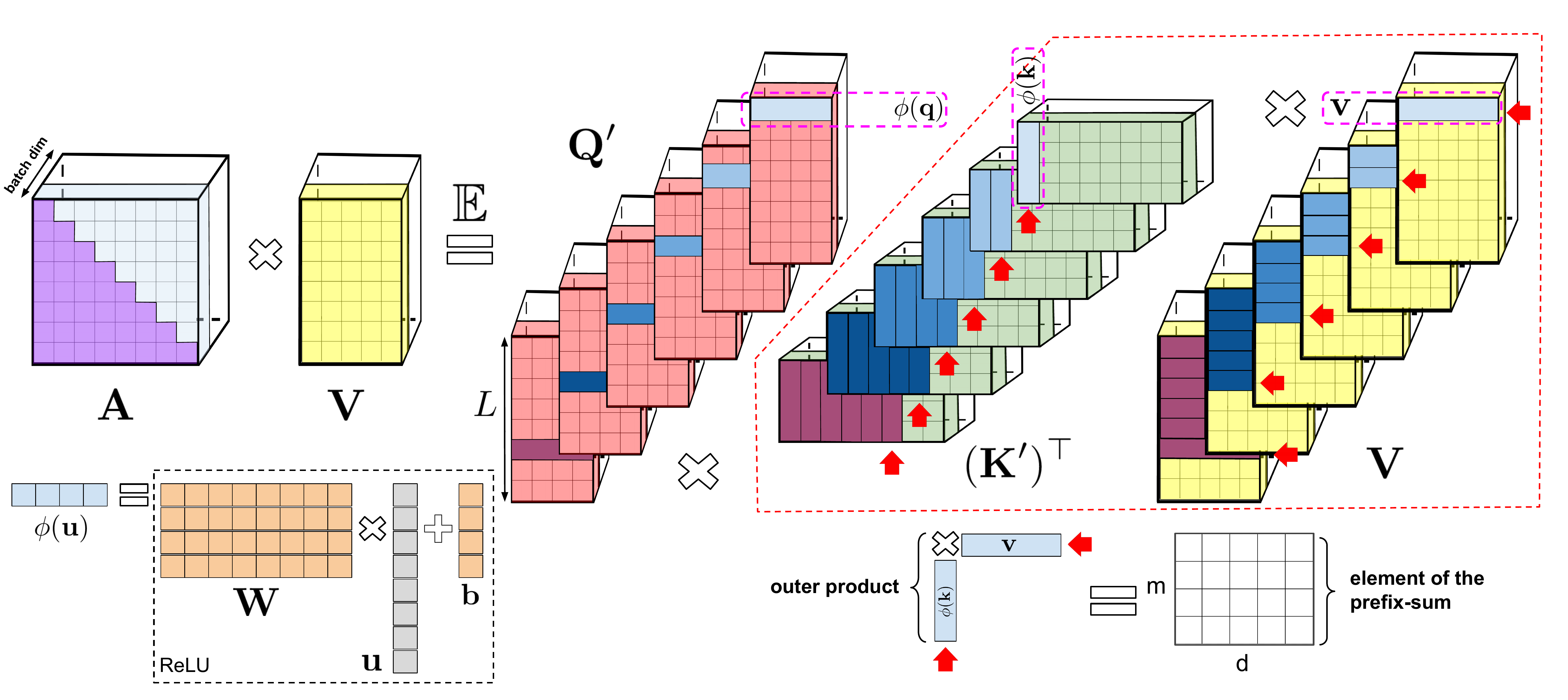}
  \caption{\small{Prefix-sum algorithm for unidirectional (causal) performer. Attention normalization is omitted. The algorithm tracks the prefix-sum: A matrix obtained by summing the outer products of kernel features corresponding to keys with value-vectors. At each given iteration of the prefix-sum algorithm, a kernel feature vector corresponding to a query is multiplied by the most recent prefix-sum (obtained by summing all outer-products corresponding to preceding tokens) to obtain new embedding. The features are obtained by applying ReLU elementwise to affinely-transformed queries/keys.}}
  \label{fig:prefix_sum}
\end{figure}

Another way to improve computational efficiency and reduce memory footprint of the conformer is to apply a recently introduced class of the linear attention techniques that avoid explicitly materializing the attention tensor, effectively replacing quadratic space and time complexity of the attention module by a more manageable linear one, and leveraging low-rank decomposition of the attention tensor. This leads to the class of models called \textit{performers} \cite{performers}. We use performers-ReLU variants from \cite{performers} inside conformers and add additional trainable affine transformation for queries/keys, as proposed in \cite{performer-finetuning}. We call the resulting conformer the \textit{RNNAttention-Performer}. The RNN-prefix in the name is motivated by the causal prefix-sum computations that unidirectional performer conducts in order to emulate causal attention, see Fig. \ref{fig:prefix_sum}. 
Our strategy is to start with regular conformer training and then replace self-attention with performer layers halfway through the training.

We ran detailed ablation studies over different performer variants in the bidirectional cascaded encoder on $\mathrm{LibriSpeech}$ data that led to the choice of the ReLu-kernel. Tested attention kernels included in particular those of the form:  
$\mathrm{K}_{f}(\mathbf{x},\mathbf{y}) = f(\mathbf{x})^{\top}f(\mathbf{y})$
where $f:\mathbb{R} \rightarrow \mathbb{R}$ is applied elementwise to kernel inputs. We benchmarked the following functions $f$: $\mathrm{ReLU}$, $\mathrm{SoftPlus}$, $\exp$, $\mathrm{ELU}$, $f(z) = z^{4}$, and settled on $\mathrm{ReLU}$ as an efficient kernel.

\subsection{Usage as a First-Pass in the Cascaded Encoder}

\begin{figure}[h!]
    \centering
    \includegraphics[width=0.6\linewidth]{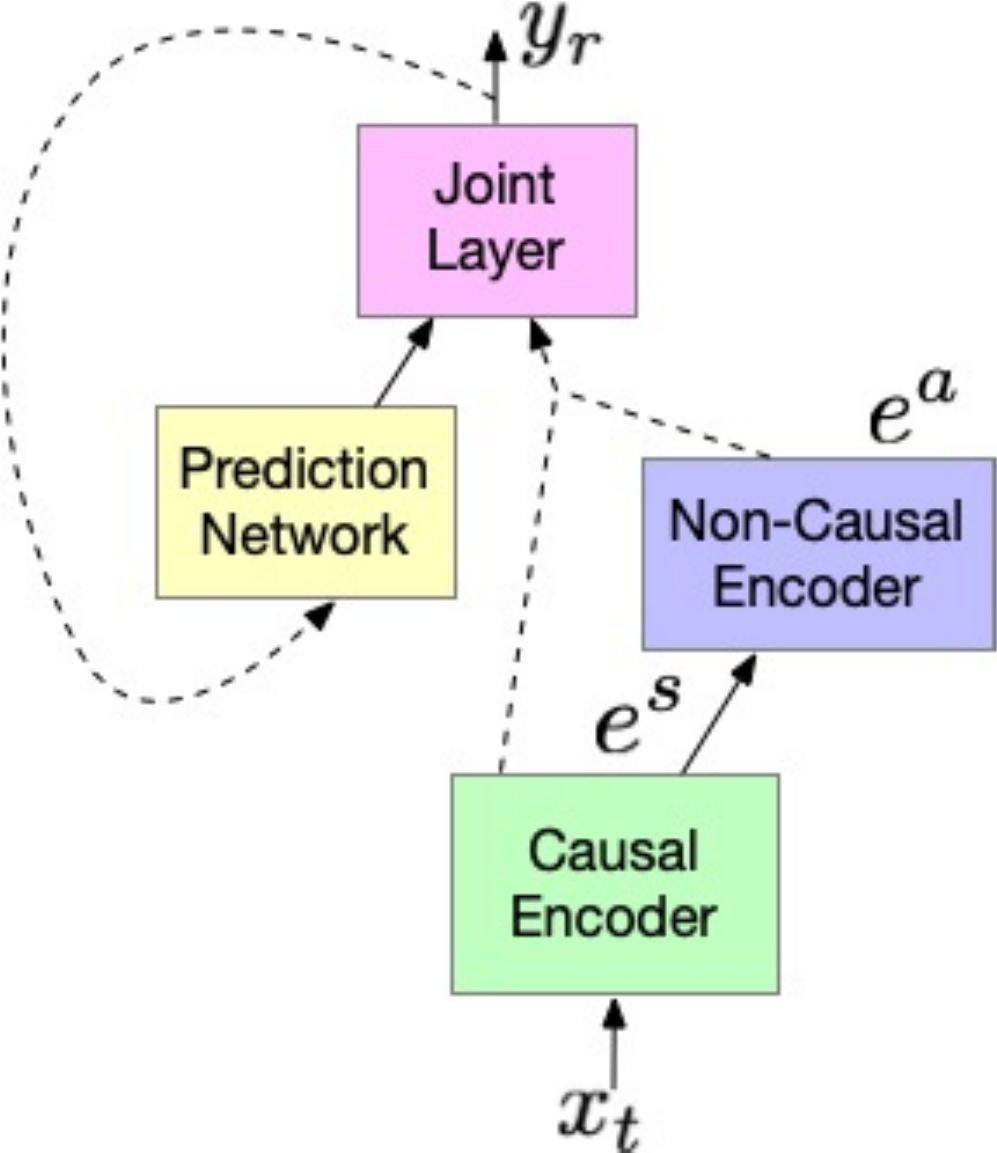}
    \caption{The cascaded encoder for joint modeling of two passes.}
    \label{fig:cascade}
\end{figure}

Meeting our size, latency and flops constraints leads to quality degradation for our optimized conformer compared to our starting baseline. In cases where the environment allows for an additional increase in flops or latency, we explore cascading a series of non-causal conformer layers on top of our encoder and running another beam search. This model, known as Cascaded Encoder \cite{arun21cascade}, is shown in Figure \ref{fig:cascade}. One motivation is that the 2nd-pass can make up for any quality degradation introduced by the small, optimized 1st-pass.

Note that the 2nd pass operates exclusively on the output of the 1st pass, without any further input from the acoustic signal. Thus, the high-latency application reutilizes the computation done by the 1st pass for the streaming setting, which contributes to overall efficiency. We also note that our optimized conformer is plugged in as a 1st-pass encoder in the cascade setting without any change in its design.

%% file: experiments.tex
\section{Experimental settings} \label{sec:experiments}

\subsection{Datasets}

As discussed in \cite{sainath20streaming}, all E2E models are trained on multidomain audio-text pairs \cite{Arun19}. All domains are anonymized and hand-transcribed, except for YouTube where the transcription is done in a semi-supervised fashion~\cite{liao2013large}. To further increase data diversity, multi-condition training (MTR) ~\cite{kim2017mtr}, random data down-sampling to 8kHz \cite{Li12} and SpecAug \cite{Park2019} are also used. Noisy data is generated at signal-noise-ratio (SNR) from 0 to 30~dB, with an average SNR of 12~dB, and with T60 times ranging from 0 to 900ms, averaging 500ms. Noise segments are sampled from YouTube and daily life noisy environmental recordings. Both 8~kHz and 16~kHz versions of the data are generated, each with equal probability, to make the model robust to varying sample rates. 

The \emph{Search} test set has around 12K Voice Search utterances with an average length of 5.5 seconds. They are anonymized, hand-transcribed, and are representative of Google's Voice Search traffic. 

\subsection{Modeling}

All models are trained on a 128D log-mel feature frontend with a 16-D one-hot domain-id vector appended to it \cite{Arun19}. Following \cite{gulati2020conformer}, the 1st-pass base conformer model uses 512D Conformer layers in the encoder. Causal convolution and left-context attention layers, with a lookback of 23 frames, are used for the Conformer layer to strictly restrict the model to use no future inputs. 8-headed self-attention is used and the convolution kernel size is 15. The encoder consists of 12 conformer blocks. Ablation studies to vary its hyperparameters will be presented in Section \ref{sec:results}. 

The RNN-T decoder comprises prediction network and a joint network with a single 640-dim FF layer. The embedding prediction network \cite{Rami21}, uses an embedding dimension of 320, and has 9M parameters. Our E2E models work with 4,096 word pieces~\cite{Schuster2012}.

The 2nd-pass cascaded encoder has 5 additional non-causal conformer layers that process a total of 900 milliseconds of future audio. Both causal and non-causal encoders feed into a shared decoder.

%% file: results.tex
\section{Results}\label{sec:results}

\subsection{Speedups Before Downsizing --- RNAP and Conv-Only}

First, Table \ref{tab:baseline_vs_rnnattn} shows our TPU-speed gains for our large 120M, once by RNNAttention-Performer and once by using converting the lowest 4 blocks to conv-only. Each method improves TPU-latency immensely with little effect on WER. Yet they bring little improvement to size and flops. Downsizing of the model is still needed for edge devices, to save space and energy, and is discussed in Section \ref{subsec:downsizing}.

\begin{table}[h!]
    \centering
    \caption{Baseline vs. RNNAttention-Performer vs. Conv-Only}
    \begin{tabular}{lcccc} \toprule
    Encoder & WER & TPU-latency (ms) & Size (M) & Flops (M) \\ \midrule
    Baseline & 6.5  & 21.8 & 120 & 248 \\ 
     RNNAP & 6.7 & 7.3 & 120 & 221 \\
     First4Conv & 6.6 & 9.5 & 113 & 223 \\ \bottomrule
    \end{tabular}
    \label{tab:baseline_vs_rnnattn}
\end{table}

\subsection{Small Conformer Model Ablation}
\label{subsec:downsizing}

\begin{table}[h!]
    \centering
    \caption{Optimized Conformer Encoder Efficient Search}
    \begin{tabular}{cccccccc}
    \toprule
    \tworow{ID} & \tworow{\FFM}  & \tworow{\NCB} & \tworow{\TB}  & \tworow{WER} & TPU & Size & Flops \\
           &  &  &   & &lat. (ms) & (M) &  (M) \\
    \midrule
     \expID{B0} & 2 & 0 & 12 & 6.5 & 21.8 & 120 & 247 \\ \hline
     \expID{E0} & 8 & 0 & 3 & 9.2 & 4.8 & 54 & 99 \\ \hline
     \expID{E1} & 6 & 1 & 4 & 8.8 & 4.6 & 54 & 96 \\ \hline
     \expID{E2} & 4 & 1 & 5 & 8.5 & 5.4 & 49 & 90 \\ \hline
     \expID{E3} & 4 & 1 & 6 & 8.2 & 5.4 & 54 & 100 \\ \hline
     \expID{E4} & 4 & 2 & 5 & 8.8 & 4.5 & 48 & 85 \\ \hline
     \expID{E5} & 4 & 3 & 7 & \textbf{7.7}  & \textbf{5.1} & \textbf{56} & \textbf{101} \\ \hline 
     \expID{E6} & 6 & 3 & 5 & 8.3 & 3.5 & 56 & 98 \\ \hline
     \expID{E7} & 2 & 4 & 8 & 8.4 & 5.3 & 42 & 73 \\ \hline
     \expID{E8} & 4 & 5 & 7 & 8.2 & 4.0 & 54 & 93 \\
     \bottomrule
    \end{tabular}
    \label{tab:grid_search}
\end{table}

Table \ref{tab:grid_search} shows an ablation study where we vary \FFM, \NCB{} and \TB, as described in Section \ref{subsection:resizing}. \expID{B0} is the baseline conformer encoder from Table \ref{tab:baseline_vs_rnnattn}. Our goal here is to hold the model size around 50M parameters and Flops around 100M, and to see how varying various hyperparameters effects both WER and TPU latency.

We have sorted the above table by $\NCB$. Notice that in \expID{E0}, if we reduce $\TB$ from 12 to 3, and increase $\FFM$, we take a large degradation in WER. In \expID{E1}, adding just 1 $\NCB$, and adjusting other parameters accordingly to keep within the size limit, improves WER. Further increasing $\TB$ in \expID{E2} and \expID{E3} improves WER again. If we continue to increase $\NCB$ to 3 at \expID{E5}, WER reaches 7.7. However, increasing  $\NCB$ or $\FFM$ further (\expID{E6-E8}), results in either too small $\TB$ or $\FFM$, which causes a quality degradation. Our best performing system, \expID{E5}, reduces the TPU latency over \expID{B0} by a factor of 4x, flops by a factor of 2.4x, and overall model size by 2.1x. It does come at a quality degradation (from 6.5\% to 7.7\%), which is addressable in less restricted environments, see Section 4.4.

Apart from strategic resizing, we also explored other techniques from the literature. First, we tried in-place distillation \cite{yu21universalasr} with the cascaded encoder in Figure \ref{fig:cascade}. Specifically, we distilled a 2nd-pass non-causal output (i.e., the teacher) to a smaller 1st-pass causal model. However, we did not see any WER improvements. Degradation was also observed when we emulated sparsity by removing connections from full-connected layers \cite{zhang2021beyond}.

\subsection{Small Model with RNNAttention-Performer}
Next, we put the optimizations with the smaller model and RNNAttention-Performer together. Table \ref{tab:final_wer_lstm_vs_conformer} shows our best model versus the conformer baseline. Our design choices, such as relatively wide FF layers, conv-only layers and RNNAttention-Performer, makes it that even though we only save 2.7X on flops, we save 6.8X on the TPU latency when using the highly parallelized computation.

\begin{table}[h!]
    \centering
    \caption{Baseline vs. Proposed Optimized Conformer Encoder}
    \begin{tabular}{lcccc} \toprule
    Encoder & WER & TPU-latency (ms) & Size (M) & Flops (M) \\ \midrule
     Baseline & 6.5  & 21.8 & 120 & 247 \\ 
     Optimized & 7.7 & 3.2 & 56 & 93 \\ \bottomrule
    \end{tabular}
    \label{tab:final_wer_lstm_vs_conformer}
\end{table}

\subsection{Cascaded Encoder}

For applications where latency/flops/size constraints can be relaxed, we explore adding additional 2nd-pass non-causal layers on top of our optimized conformer, reusing its same design as part of a higher-quality model. Table \ref{tab:2ndpass} shows the WER for both passes of the cascaded model. We compare our setup with a baseline \cite{sainath2021cascadedlm} that has a large first-pass and a small second-pass. Both settings achieve the same second-pass WER, therefore the first pass encoder does not need to be large in order to serve the 2nd-pass with an informative encoded representation. Thus, for a given a total size of a cascade model, the bulk of parameters can be allocated to the second pass, allowing the first-pass encoder to also be deployed in a small on-device model. Note that the first-pass WER stays at 7.7\% the same as when it was trained alone (Table \ref{tab:final_wer_lstm_vs_conformer}). Thus, even after the joint training, it can still function as a standalone encoder for edge devices.

\begin{table}[h!]
    \centering
    \caption{WER of 1st+2nd-pass cascade models}
    \begin{tabular}{lccc} \toprule
    1st-pass Encoder & Pass & WER & Size (M)\\ \midrule
    \tworow{Large Baseline}  & 1st & 6.2 & 120  \\
       & 2nd & 5.8  & 50 \\ \midrule
    \tworow{Small Optimized} & 1st & 7.7 & 56  \\
       & 2nd & 5.8 & 100 \\ \bottomrule   
    \end{tabular}
    \label{tab:2ndpass}
\end{table}

%% file: conclusions.tex
\section{Conclusions}
\label{sec:conclusion}

In this work, we designed a causal, streaming conformer that meets practical size, latency and flop criteria, while still being informative enough for a non-causal large second-pass. We addressed the issue of slow inference of self-attention layers by showing that the first 3 can be removed and the rest can be replaced with performer layers. A careful search was conducted to choose hyperparameters that balance speed and size. Overall, the size is halved and inference runs 6x faster on TPU. Though this leads to a degradation in WER, if additional compute time is permissible, a cascaded 2nd-pass encoder brings the quality back up to the baseline level.